# High-Quality Pluralistic Image Completion via Code Shared VQGAN

CHUANXIA ZHENG, Monash University, Australia
GUOXIAN SONG and TAT-JEN CHAM, Nanyang Technological University, Singapore
JIANFEI CAI and DINH PHUNG, Monash University, Australia
LINJIE LUO, ByteDance Inc, USA

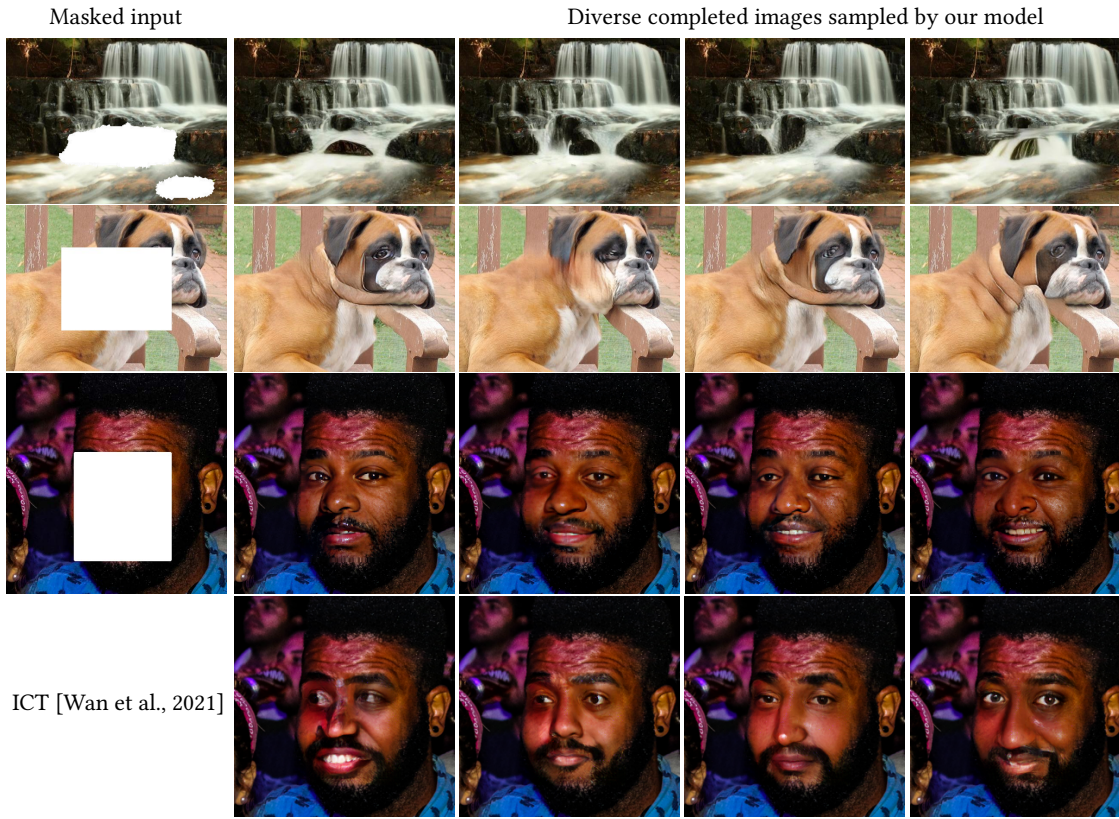

Fig. 1. Our method produces high-quality and diverse image completion results on input masked images from various datasets (from top to bottom: Places2 [Zhou et al., 2018], ImageNet [Russakovsky et al., 2015] and FFHQ [Karras et al., 2019] respectively). Compared to ICT [Wan et al., 2021] (final row), our method not only generates more visually plausible results with fewer artifacts, but also runs much faster with more than 100x speed-up in inference time (averaged 83ms on a single GPU and 1.68s on a CPU). More diverse results are represented as **animations** in the last column.

PICNet pioneered the generation of multiple and diverse results for image completion task, but it required a careful balance between $\mathcal{KL}$ loss (diversity) and reconstruction loss (quality). Separately, iGPT-based architecture has been employed to infer distributions in a discrete space derived from a pixel-level pre-clustered palette, which however cannot generate high-quality results directly. In this work, we present a novel framework for pluralistic image completion that can achieve both high quality and diversity at much faster inference speed. The core of our design lies in a simple yet effective code sharing mechanism that leads to a very compact yet expressive image representation in a discrete latent domain. The compactness and the richness of the representation further facilitate the subsequent deployment of a transformer to effectively learn how to composite and complete a masked image at the discrete code domain. Based on the global context well-captured by the transformer and the available visual regions, we are able to sample all tokens simultaneously, which is completely different from the prevailing autoregressive approach of iGPT-based works, and leads to more than 100× faster inference speed. Experiments show that our framework is able to learn semantically-rich discrete codes efficiently and robustly, resulting in much better image reconstruction quality. Our diverse image completion framework significantly outperforms the state-of-the-art both quantitatively and qualitatively on multiple benchmark datasets.

Authors' addresses: Chuanxia Zheng, chuanxia.zheng@monash.edu, Monash University, Australia; Guoxian Song, guoxian001@e.ntu.edu.sg; Tat-Jen Cham, ASTJCham@ntu.edu.sg, Nanyang Technological University, Singapore; Jianfei Cai, jianfei.cai@monash.edu; Dinh Phung, dinh.phung@monash.edu, Monash University, Australia; Linjie Luo, linjie.luo@gmail.com, ByteDance Inc, USA.

# 1 INTRODUCTION

Image completion, also named "inpainting" [Bertalmio et al., 2000], is a task of filling masked regions with alternative realistic content seamlessly. Applications include restoring damaged paintings [Bertalmio et al., 2000], removing unwanted objects [Criminisi et al., 2003], generating new content for occluded regions [Zheng et al., 2021b], and freely editing content in an image [Nazeri et al., 2019].

To infer plausible content, many learning-based approaches [Iizuka et al., 2017, Nazeri et al., 2019, Pathak et al., 2016, Yu et al., 2018, Zheng et al., 2022] have been proposed in the last few years. These methods generate new content by learning statistical information from large datasets. However, most of them provide *only a single solution* to a given masked input, despite the multi-modal nature of the problem. PICNet [Zheng et al., 2019] is a pioneering effort that aimed to generate *multiple* and *diverse* plausible results. The basic idea of PICNet is to learn a continuous distribution of solutions that is compatible with the visible pixels, and then sample from it to achieve diverse results. While it carefully sought a balance between $\mathcal{KL}$ loss and reconstruction loss in a conditional variational encoder (CVAE) setting, its conventional formulation and architecture resulted in limited diversity and quality.

Inspired by iGPT [Chen et al., 2020], some recent efforts [Wan et al., 2021, Yu et al., 2021] have been made in directly predicting appropriate tokens' possibility in discrete latent space, which is more robust to posterior collapse. However, these works use a *pre-clustered* palette at the *pixel level*, leading to reduced image quality. Although super-resolution techniques can be applied to improve image quality, the performance is limited by the upsampling scale (4×), which cannot generalize to arbitrary scenes. While the recent VQVAEv2 [Razavi et al., 2019] can generate high-quality images using learned discrete features, they use multi-scale indices, i.e. $32 \times 32$ and $64 \times 64$, resulting in very long token sequences that cannot be easily handled with a transformer architecture. Although convolutional neural networks (CNN) may be deployed to model long token sequences, the generated content may not be globally consistent due to the inherently local nature of CNNs.

Therefore, in this work we propose a novel discrete code-sharing representation that can represent an image as *discrete* code indices in a very low resolution, *e.g.* $16 \times 16$. Our key insight is that *learning to represent a high-dimensional image as a composition of low-dimension building blocks naturally leads to efficient and effective image modeling that can greatly facilitate downstream tasks like image completion*. In particular, instead of quantizing each spatial grid feature as a codebook entry, we propose to subdivide each grid feature along its channel dimension into multiple chunks, and then use a code entry to represent each chunk. More importantly, the chunk codes come from a shared codebook. In this way, an image can be considered as a unique composition of shared building blocks, *i.e.* shared codes from a common codebook, along both spatial and channel dimensions. Such a code-sharing mechanism facilitates the *disentanglement and recomposition of shared attributes in the discrete latent domain*, and results in a simple, efficient yet effective representation that achieves higher image quality than state-of-the-art methods using the same $16 \times 16$ scale indices (Figs. 2 (b) and (c)).

With this practical and expressive sequencing approach, we can then leverage a transformer architecture to predict the appropriate composition of a code sequence, exploiting the ability of the transformer to capture the global context relationship in every layer. Specifically, a weighted bidirectional attention module is introduced to enforce the network to focus more on visible regions. Furthermore, thanks to the well captured global context relationship in the transformer and the available visible regions in completion, we are able to simultaneously sample all tokens at one time, which is completely different to the prevailing autoregressive approach for sequence generation. This leads to more than 100× faster inference compared to the latest autoregression-dependent methods [Wan et al., 2021, Yu et al., 2021], as we avoid a recurrent formulation. Finally, a refinement network is employed to further improve image quality and allows the whole framework to handle images of arbitrary sizes.

We comprehensively evaluated and compared our approach with existing state-of-the-art methods on a large variety of scenes, which are degraded by various regular or free-form irregular masks. The extensive results show that our approach consistently outperformed previous methods both quantitatively and qualitatively. We also show various image editing applications of our model to a range of tasks, such as object removal and content generation.

To summarize, the main contributions of this paper are as follows:

- We propose a novel code-sharing mechanism in the discrete latent domain that facilitates a very compact image presentation while still able to reconstruct images at high quality.
- With our proposed compact yet expressive token representation, we adopt a transformer architecture to capture global context dependencies, with customized new components including a weighted bidirectional attention scheme and a simultaneously sampling strategy, which greatly improves the inference speed.
- We also develop an interactive system, supporting various image editing applications such as object removal and content generation, where users can easily edit images via masks or sketched input.

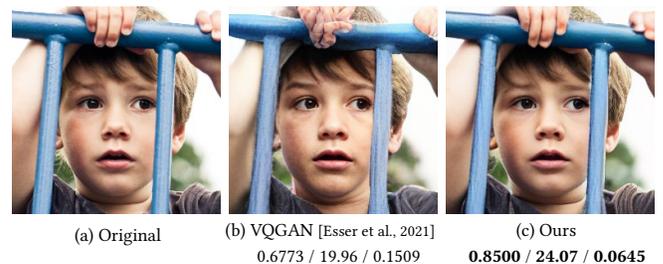

(a) Original     (b) VQGAN [Esser et al., 2021]     (c) Ours

0.6773 / 19.96 / 0.1509      **0.8500** / **24.07** / **0.0645**

Fig. 2. Reconstructed results of different methods using discrete latent codes. The image quality is evaluated using SSIMs, PSNR and LPIPS. (a) Input image. (b) VQGAN [Esser et al., 2021] applied $16 \times 16$ scale feature for VQ representation. While the image quality is high, some details are lost (*e.g.* the hand and hair for the child). (c) In contrast, our proposed code-shared method, by quantizing different channel chunks into different codes (still in $16 \times 16$ resolution), results in a much better reconstruction quality, which helps the image completion task significantly (as shown in Fig. 10).

## 2 RELATED WORK

A variety of image completion models have previously been proposed. These approaches either utilize the visible information from within the image (**intra-image**) [Bertalmio et al., 2000, 2003], or learn the statistical information from a large dataset (**inter-image**) [Hays and Efros, 2007, Iizuka et al., 2017].

*Intra-Image Completion.* Classical image completion works, also known as "image *inpainting*" [Bertalmio et al., 2000], directly propagate, copy or realign visible pixels to missing regions, based on the assumption that missing regions should be filled with similar appearances as those in visible regions. One category of the *intra*-image completion methods is the diffusion-based image synthesis [Ballester et al., 2001, Bertalmio et al., 2000, 2003, Levin et al., 2003]. These approaches propagate the locally surrounded visible pixels to the missing regions, resulting in smooth results, yet work only on some small and narrow holes. In contrast, patch-based approaches [Barnes et al., 2009, Criminisi et al., 2003, 2004, Jia and Tang, 2004], are able to deal with larger and more complex holes. They copy and realign the pixels from visible regions to missing regions, by analyzing and parsing the low-level features in multiple patches. These approaches produce texture-consistent images. However, they only utilize information within a single image, and thus they are *not* able to generate semantically new content for large holes.

*Inter-Image Completion.* To ameliorate the problem of generating new content for large holes, *inter*-image completion approaches borrow statistical information from a large dataset. In particular, Hays and Efros [2007] applied the image retrieval method to directly search the most similar image for the masked one from a huge dataset and then filled new content by cutting the corresponding regions from the retrieval image and pasting them into the missing regions. However, it requires the dataset to be large enough to contain an image similar to the arbitrary masked input image, which might be hard to be met. Even if we can collect such a huge dataset, the searching cost will be expensive for each masked image.

Recently, learning-based approaches have become prevailing for image completion. Initially, Köhler et al. [2014] and Ren et al. [2015] introduced Convolutional Neural Networks (CNNs) [LeCun et al., 1998] for the image completion task. However, they can only address small and thin missing regions. In contrast, Pathak et al. [2016] made CNNs be able to perform more complicated image completion with 64x64-sized holes, by utilizing the emerging Generative Adversarial Networks (GANs) [Goodfellow et al., 2014]. Then, Iizuka et al. [2017] built upon the Context Encoder (CE) [Pathak et al., 2016] for arbitrary regular masks, by combining Globally and Locally (GL) discriminators for the adversarial learning loss. More recently, Liu et al. [2018] introduced "partial convolution" for free-form irregular mask image completion, and a corresponding irregular mask dataset was provided. On the other hand, various additional information was explored for semantical image completion. In particular, Li et al. [2017] generated semantically-consistent completed images by applying an additional face parsing loss. Song et al. [2018b] utilized the completed semantic map as guidance for RGB image completion. This was also extended to various cases in which sketches, colors, and semantics are used in image completion. FaceShop [Portenier et al., 2018] and SC-FEGAN [Jo and Park, 2019] directly input the sketch and color for high-resolution face editing. DeepFillv2 [Yu et al., 2019] and EdgeConnect [Nazeri et al., 2019] applied edges for the general image completion problem. While these approaches learn the statistical information from a large dataset and then infer reasonable content for a masked input image, their completed appearances may *not* be consistent with the original visible pixels, particularly for large missing regions.

*Intra- and Inter-Image Completion.* To generate semantically reasonable content and visually consistent appearance that smoothly match the original visible regions, a common strategy is to combine *intra*- and *inter*- image completion methods. For instance, Yang et al. [2017] applied multiscale neural patch synthesis for high resolution image completion, which generated high-frequency details by copying and realigning patches based on the mid-layer features. Inspired by PatchMatch [Barnes et al., 2009], Yu et al. [2018] introduced a Contextual Attention (CA) module, in which a feature-based Patch-Match algorithm was proposed to copy the similar features from visible regions to missing regions. This is followed by a series of works [Li et al., 2020, Liao et al., 2021, Song et al., 2018a, Suin et al., 2021, Yan et al., 2018, Yi et al., 2020, Zeng et al., 2020, 2021, Zheng et al., 2022]. In particular, Yan et al. [2018] and Song et al. [2018a] extended PatchMatch ideas on feature domain. Then, Yi et al. [2020] proposed contextual residual aggregation for high-resolution (8K) image completion. Zeng et al. [2020] and Li et al. [2020] applied the attention module in a recurrent way to optimize the appearance in multiple stages. Suin et al. [2021] further applied auxiliary attention blocks in multiple feature scales. More recently, Zheng et al. [2022] built their model upon the transformer architecture [Vaswani et al., 2017] and designed an automatically attention-aware layers for high-quality image completion. However, all these approaches generate only one single result for one masked image.

*Pluralistic Image Completion.* Zheng et al. [2019, 2021a] first proposed "pluralistic image completion (PIC)" task, which aims at generating multiple and diverse results for each masked input image. To achieve this goal, a dual pipeline framework is introduced, where one path is built for reconstruction and the other is for producing diverse results. Similar with [Zheng et al., 2019], Zhao et al. [2020] and Liu et al. [2021] optimized the $\mathcal{KL}$-divergence between encoded features and $\mathcal{N}(\mathbf{0}, \mathbf{I})$, and then sampled from the given distribution to generate diverse results. Although these approaches are able to provide more diversified results, their image quality can not always be guaranteed due to the variational training. Inspired by the impressive performance of image Generative Pre-Training (iGPT) [Chen et al., 2020], Wan et al. [2021] and Yu et al. [2021] directly applied this architecture for pluralistic image completion by computing the log-likelihood score for each token in a discrete space, instead of sampling from a continuous space, *i.e.* $\mathcal{N}(\mathbf{0}, \mathbf{I})$. However, their discrete space is a pre-clustered palette on pixel-level, in which images are downsampled to a small resolution, *e.g.* $32 \times 32$. This may not impact the image classification task [Chen et al., 2020, Torralba et al., 2008], but the generated results are of low-quality and unimpressive visual appearance.

In the last few years, vector quantization (VQ) approaches [Esser et al., 2021, Razavi et al., 2019, van den Oord et al., 2017] have been applied to image generation and completion, which achieve excellent results using a learned codebook in the feature domain. Inspired by these methods, Peng et al. [2021] directly applied the VQ-based learning pipeline for the image completion task. All these approaches generate new content in an autoregressive way, which can ensure structural consistency but at the cost of taking quite a long time during the inference. We avoid this issue by directly sampling all invisible contents at one time, as the multiple self-attention mechanisms in our transformer architecture have successfully modelled the global context relationships within an image.

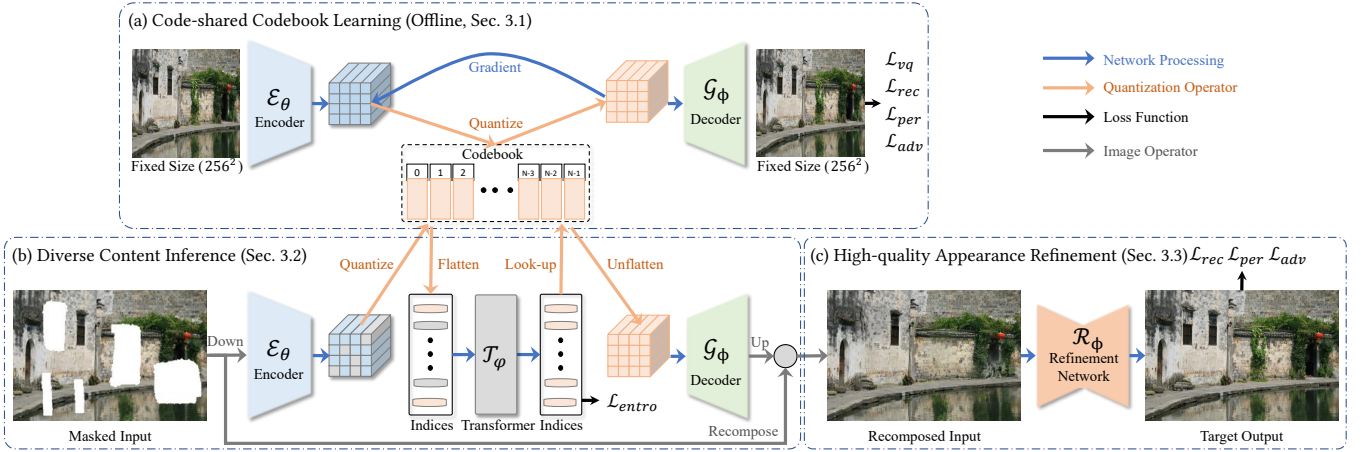

Fig. 3. An overview of our pipeline for high-quality pluralistic image completion. (a) Our approach first learns a better codebook of context-rich visual parts, by using a code-shared strategy. (b) A transformer architecture is then applied to globally infer the composition of original embedded indices for the masked tokens. (c) Finally, we sample the top $\mathcal{K}$ results, and then merge the completed coarse result with the original high-quality image. A refinement network is designed to ameliorate the visual gap between visible and generated masked regions. Note that only the bottom pipeline is used during inference, while the top pipeline is for learning the codebook with the shared encoder and decoder offline.

## 3 METHODOLOGY

Given a single masked image $x_m$, our goal is to generate high-quality completed images from the partially visible regions. We opt to go beyond a single "best" solution to deal with *multiple* and *diverse* results. This requires capturing the global statistics of complex image structures to *infer* diverse reasonable contents, as well as producing fine details and texture information with visually consistent appearance between visible and generated regions.

To sample multiple and diverse results, we learn to better predict the tokens' distribution in a *discrete* space, rather than mapping the whole dataset to a given continuous distribution, *e.g.* $\mathcal{N}(\mathbf{0}, \mathbf{I})$ in previous works [Liu et al., 2021, Zhao et al., 2020, Zheng et al., 2019]. As shown in Fig. 3, our pipeline consists of three major stages during training. We use a transformer architecture (Fig. 3 (b)) to efficiently infer missing tokens' distribution. This architecture is similar to the existing image completion works [Wan et al., 2021, Yu et al., 2021], except that here we learn a better codebook (Fig. 3 (a)) to embed a high-resolution image as a composition of code indices with lower resolution on *feature-level*, rather than directly downsampling an image and embedding with a pre-clustered palette on *pixel-level*. With such an effective learnable codebook, both masked input image $x_m$ and ground truth target image $x_{gt}$ are respectively embedded and flattened into index sequences $\mathbf{s}_m$ and $\mathbf{s}_{gt}$, and a transformer is then trained to infer the degraded indices by exploiting the global interrelations within a sequence of tractable length (Fig. 3 (b)). Finally, the refinement network is designed to further enhance the resolution and visual quality of completed images (Fig. 3 (c)).

Over the following sections, we will introduce a novel code-shared codebook learning scheme for more efficient and structured discrete representation (Sec. 3.1), and a transformer architecture with multiple weighted bidirectional self-attention modules to effectively infer diverse content for missing regions (Sec. 3.2). In Sec. 3.3, an attention-based refinement network is employed to further ensure visually consistent appearance.

### 3.1 Code-shared Codebook Learning

*Quantization.* Our feature embedding approach is based on the latest vector quantization methods [van den Oord et al., 2017]. Given an image $x \in \mathbb{R}^{H \times W \times 3}$, it will be represented as a spatial composition of codebook entries $z_q \in \mathbb{R}^{h \times w \times n_z}$, where a sequence $\mathbf{s}$ of $h \cdot w$ indices of these entries can be used as an equivalent representation. In particular, an image is reconstructed by

$$\hat{x} = \mathcal{G}_\phi(\mathbf{z}_q) = \mathcal{G}_\phi(\mathbf{q}(\mathcal{E}_\theta(x))), \quad (1)$$

where the encoder $\mathcal{E}_\theta$ embeds an image $x$ into a high-dimensional low-resolution feature $\hat{\mathbf{z}} \in \mathbb{R}^{h \times w \times n_z}$, with the same dimensionality as the codebook entries $\mathbf{z}_q$, and the decoder $\mathcal{G}_\phi$ reverse transfers the quantized feature $\mathbf{z}_q = \mathbf{q}(\hat{\mathbf{z}})$ back to the image domain. The quantized operator $\mathbf{q}(\cdot)$ is performed by matching the closest entry $z_k$ in the codebook for each spatial grid feature $\hat{z}_{ij}$ in $\hat{\mathbf{z}}$:

$$\mathbf{z}_q = \mathbf{q}(\hat{\mathbf{z}}) = \arg \min_{z_k \in \mathbb{Z}} \|\hat{z}_{ij} - z_k\|. \quad (2)$$

*Loss Functions.* The optimization target is conceptually similar to that for conventional generators, except here *the discrete codebook*

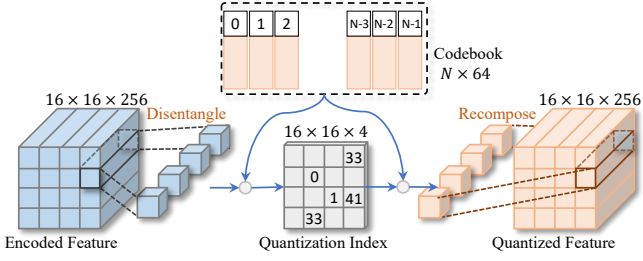

Fig. 4. Illustration of code-shared codebook learning. Unlike existing methods [Esser et al., 2021, van den Oord et al., 2017] that quantize the whole feature, we first separate each feature into multiple groups (e.g. 4 or 8), and then quantize these sub-features using the learned codebook. By do this, we aim to disentangle attributes in feature channels which can be reused or recomposed for different spatial features.

$\mathbb{Z} = \{z_k\}_{k=1}^{K}$ *is also updated*, instead of only optimizing the network parameters $\theta$ and $\phi$ in the encoder $\mathcal{E}_\theta$ and the decoder $\mathcal{G}_\phi$. Following VQ-GAN [Esser et al., 2021], multiple loss functions are applied to learn a perceptually rich codebook:

$$\mathcal{L}_{VQ}(\mathcal{E}_\theta, \mathcal{G}_\phi, \mathbb{Z}) = \mathcal{L}_{rec} + \mathcal{L}_{per} + \mathcal{L}_{vq} + \mathcal{L}_{adv}. \quad (3)$$

The first and second terms are the data term for reconstruction and perceptual losses, respectively, which is primarily for optimizing the encoder $\mathcal{E}_\theta$ and the decoder $\mathcal{G}_\phi$. Specifically, we use $\mathcal{L}_1$ loss for $\mathcal{L}_{rec}$ to measure the pixel-level reconstruction loss between input image $x$ and generated output $\hat{x}$:

$$\mathcal{L}_{rec} = \|x - \hat{x}\|. \quad (4)$$

In addition, the learned perceptual image patch similarity (LPIPS) [Zhang et al., 2018] is utilized to measure the feature-level similarity:

$$\mathcal{L}_{per} = \|\Phi_n(x) - \Phi_n(\hat{x})\|, \quad (5)$$

where $\Phi_n$ is the activation map of the $n$th selected layer in LPIPS. The third term in (3) measures the distance between the encoded feature $\hat{z}$ and the quantized feature $\mathbf{z}_q$, given by

$$\mathcal{L}_{vq} = \|\text{sg}[\mathcal{E}_\theta(x)] - \mathbf{z}_q\|^2 + \beta\|\mathcal{E}_\theta(x) - \text{sg}[\mathbf{z}_q]\|^2, \quad (6)$$

where "sg" stands for the stop-gradient operator [van den Oord et al., 2017], which forces the corresponding operand to be a non-updating constant during backpropagation. Therefore, this loss encourages the learned quantized features $\mathbf{z}_q$ to move towards the encoded features $\mathcal{E}_\theta(x)$ in the first part of (6), while concurrently making the encoded features commit to the codebook space via the second part of (6). More details can be found in the VQ-VAE paper [van den Oord et al., 2017]. To further improve image quality, an adversarial training loss [Goodfellow et al., 2014] is introduced in VQ-GAN [Esser et al., 2021]:

$$\mathcal{L}_{adv} = \arg\min_{\mathcal{E}_\theta, \mathcal{G}_\phi, \mathbb{Z}} \max_{\mathcal{D}} \mathbb{E}_{x \sim p(x)}[\log D(x) + \log(1 - D(\hat{x}))], \quad (7)$$

where $\hat{x}$ is the output image as in (1). In practice, we optimize the hinge version of this adversarial loss [Miyato et al., 2018].

*Code-shared Codebook.* Our broad conjecture here is that *the feature in a single spatial location may consist of multiple attributes that should be disentangled, and features in different locations can share some common attributes.* Motivated by this, we propose to chunk each grid feature into multiple tablets or segments, and then embed each of them into a code in the codebook. Note that this is very different from existing VQ-based generators, which embed each spatial grid feature entirely to its closest codebook entry. Our *key novel insight* is that doing so will allow the network to learn to decouple the high-level semantics and attribute segments along the channel dimension, while at the same time supporting a rich combinatorial representation from the flexible recomposition of these segments, yet without increasing the number of learnable weights in the encoder and decoder.

Our proposed pipeline is illustrated in Fig. 4. Given an encoded feature $\hat{z} \in \mathbb{R}^{h \times w \times n}$, we directly subdivide it along the channel dimension into multiple chunks, *i.e.* $\hat{z}_{ij} = \{\hat{z}_{ij}^{(1)}, \hat{z}_{ij}^{(2)}, \hat{z}_{ij}^{(3)}, \hat{z}_{ij}^{(4)}\}$, $\hat{z}_{ij}^{(k)} \in \mathbb{R}^{n/4}$. Then, each chunk is quantized to its closest entry $z_k$ in the codebook using (2), where $z_k \in \mathbb{R}^{n/4}$ and the equivalent index representation becomes 4-channel. The quantized features can be recomposed and reused for other spatial positions. Note that, although our quantized feature indices have multiple channels, the length of the index sequence $\mathbf{s}$ is still $h \cdot w$, as used for the subsequent transformer training.

### 3.2 Diverse Content Inference

*VQ-Transformer.* To infer diverse plausible content for missing regions, we need to train a network to effectively capture the global context information. In particular, with the above trained codebook $\mathbb{Z}$, an image can now be represented as a set of code indices in low resolution, *i.e.* $h \times w = 16 \times 16$, which can be further flattened into a sequence $\mathbf{s} = [s^1; s^2; \ldots; s^N]$, where $N = 16 \times 16$. With this practical sequence length of $N$, we can apply the highly expressive transformer encoder [Vaswani et al., 2017] for image completion, by modeling the global context information in every attention layer. More precisely, given an index sequence $\mathbf{s}$, we first project each of its elements into a $C$-dimensional feature to get the sequence $\mathbf{E} = [\mathbf{e}^1; \mathbf{e}^2; \ldots; \mathbf{e}^N]$ through a learnable embedding. Note that due to our code-sharing mechanism, each sequence symbol $s^n$ has multiple channels, *i.e.* 4 as shown in Fig. 4. We obtain each embedded feature $\mathbf{e}^n$ by an implementation trick whereby we embed each channel of $s^n$ (corresponding to a code index) via the same learnable embedding and then concatenate them together.

In this way, the block in the transformer encoder for the embedded sequence $\mathbf{E}$ can be written as:

$$\mathbf{E}_0 = [\mathbf{e}^1; \mathbf{e}^2; \ldots; \mathbf{e}^N] + \mathbf{E}_{pos}, \quad (8)$$

$$\mathbf{E}'_\ell = \text{MSA}(\text{LN}(\mathbf{E}_{\ell-1})) + \mathbf{E}_{\ell-1}, \quad (9)$$

$$\mathbf{E}_\ell = \text{MLP}(\text{LN}(\mathbf{E}'_\ell)) + \mathbf{E}'_\ell, \quad (10)$$

where $\mathbf{E} \in \mathbb{R}^{N \times C}$ is a sequence of $N$ tokens each with $C$ channels (we use $C$=64), $\mathbf{E}_{pos}$ is the position embedding, and MSA, MLP, LN are respectively the `Multi-head Self-Attention`, `Multi-Layer Perception`, and `LayerNorm`. More specifically, MSA is responsible for capturing global interrelations within a sequence through the

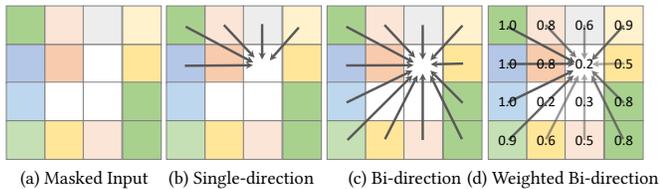

(a) Masked Input  (b) Single-direction  (c) Bi-direction  (d) Weighted Bi-direction

Fig. 5. Weighted bidirectional self-attention mechanism in our transformer-encoder. We not only use the bidirectional attention to perceive the global visible information from both sides, but also assign a weight to explicitly indicate the confidence in each embedded token, by calculating its visible pixel ratio.

pairwise similarity:

$$[\mathbf{q}, \mathbf{k}, \mathbf{v}] = \mathbf{W}_{qkv}\mathbf{E}_\ell, \quad (11)$$

$$\mathrm{SA}(\mathbf{E}_\ell) = \mathrm{softmax}(\mathbf{q}\mathbf{k}^\top/\sqrt{C_h})\mathbf{v}, \quad (12)$$

$$\mathrm{MSA}(\mathbf{E}_\ell) = [\mathrm{SA}_1; \dots; \mathrm{SA}_h], \quad (13)$$

where $\mathbf{W}_{qkv} \in \mathbb{R}^{C \times 3C_h}$ is a learnable linear projection to transfer feature $\mathbf{E}_\ell$ as query $\mathbf{q}$, key $\mathbf{k}$ and value $\mathbf{v}$, and $h$ is the number of heads. MLP is then applied to transform merged features of MSA, and their outputs are further processed with a non-linear projection using LN for the next block.

*Weighted Bi-directional Self-attention.* One common aspect of many existing transformer-based generators [Brown et al., 2020, Chen et al., 2020, Radford and Narasimhan, 2018, Radford et al., 2019] is that their attention module is always single-direction, which only calculates long-range dependencies with the previous tokens (Fig. 5 (b)). While this may not impact the sequence generation task, it has a major drawback for the image completion task, i.e the completed results often have inconsistent content and appearance with the original bottom-right visible pixels, as shown in Fig. 10 (b). Inspired by BERT [Devlin et al., 2018], ICT [Wan et al., 2021] and BAT-Fill [Yu et al., 2021] mitigated this issues by using a bidirectional attention to capture global context information from all positions (shown in Fig. 5 (c)), where they directly denoted the degraded pixels as [MASK] tokens. However, their frameworks require the use of autoregressive sampling to get high quality results. Sequentially generating tokens one-by-one results in a long inference time (average 2min30s for ICT [Wan et al., 2021] on an NVIDIA 3090 GPU for a 2min masked image), thus limiting their applications.

To achieve high quality results without the need for slow autoregressive sampling, besides using our much more effective code-shared codebook representation, we follow a recent effort [Zheng et al., 2022] in utilizing a weighted bidirectional attention module to model the global context information based on the visibility factor. In particular, the input masked image is passed through a pre-trained encoder to produce the embedded feature, as well as the initial weight (as shown in Fig. 5 (d)). This weight is obtained by calculating the percentage of visible pixels in the corresponding downsampled scale, *e.g.* $64/(16 \times 16) = 0.25$ means 64 pixels in a $16 \times 16$ patch are visible, *which explicitly indicates the significance of embedded features in different spatial positions*. Then, the feature and weight are passed through the transformer with the weighted bidirectional attention, where the original attention score $\mathrm{SA}(\mathbf{E}_\ell)$ in Eq. (13) will be scaled by the weight $\mathbf{w}_\ell$. The weight $\mathbf{w} \in (0.02, 1]$ is updated as $\sqrt{\mathbf{w}_{\ell-1}} \to \mathbf{w}_\ell$ after each transformer block. Note that although the weighted bidirectional attention has also been used in TFill [Zheng et al., 2022], unlike TFill, our embedded features are quantized and the corresponding sequence $\mathbf{s}$ is embedded as tokens. More importantly, the final layer in our transformer is used to predict a per-chunk (*i.e.* per-element and per-channel) distribution over $K$ indices in the learned codebook for diverse sampling, instead of directly projecting the continuous features to a decoder.

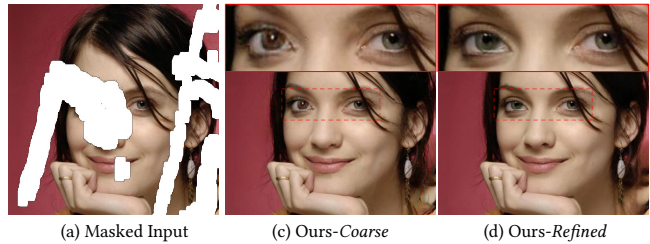

(a) Masked Input  (c) Ours-*Coarse*  (d) Ours-*Refined*

Fig. 6. Examples of coarse and refined completion results. While the coarse results contain semantically reasonable content and visually plausible appearance, they are not always consistent to the original visible appearance (*e.g.* the colors of eyes). The refinement can ameliorate the issue by shuffling high-quality feature from visible to generated regions.

*Loss Function.* Given indices $\mathbf{s}_m$, quantized from the masked input image $x_m$, the transformer $\mathcal{T}_\psi$ learns to predict the distribution of possible indices, *i.e.* $p(\mathbf{s}|\mathbf{s}_m)$, with the target indices $\mathbf{s}_{gt}$ that is quantized from the original ground truth image $x$. The learning is guided by minimizing the negative log-likelihood of the data representations:

$$\mathcal{L}(\mathcal{T}_\psi) = \mathcal{L}_{entro}(\mathbf{s}_{gt}, \mathcal{T}_\psi(\mathbf{s}_m)) = \mathbb{E}_{x \sim p(x)}[-\log p(\mathbf{s}|\mathbf{s}_m)]. \quad (14)$$

*Sampling Strategy.* Given the generated discrete distribution, we can sample the indices at different chunks (*i.e.* at each position and channel). Specifically, the transformer predicts the probability of $K$ codebook entries for all chunks, and then we sample tokens based on their confidence scores. As opposed to prior works, we observe that *independently sampling all positions can also produce good results*, which dramatically reduces the inference time. This is because the global dependencies have been modeled in every transformer layer through the attention module, and the generated tokens are constrained by visible tokens.

### 3.3 High-quality Appearance Refinement

Through the codebook learning and the global context modeling, the proposed coarse pipeline (Fig. 3 (b)) can provide *multiple* and *diverse* results for a given masked image. However, the coarse pipeline has several limitations. First, while our code-shared method significantly improves the performance of the codebook representation, the quantization still discards some visual details. Second, the diverse content is inferred using embedded low-resolution indices. These factors may not result in fully consistent appearance between visible and completed pixels. Furthermore, the coarse pipeline only works at a fixed resolution, *i.e.* $256 \times 256$, due to the fixed-length position embedding in the transformer.

Table 1. Quantitative comparisons on Places2 [Zhou et al., 2018] with free-form masks [Liu et al., 2018]. Without bells and whistles, the proposed method outperforms existing learning-based models on most metrics, especially for the feature-level metrics. Following established works, results are mainly reported on 256 × 256 resolution, except that our refined results are reported on 512 × 512 resolution.

| Mask Ratio | PSNR ↑ | | | SSIM ↑ | | | LPIPS ↓ | | | FID ↓ | | |
|---|---|---|---|---|---|---|---|---|---|---|---|---|
| | 20-30% | 30-40% | 40-50% | 20-30% | 30-40% | 40-50% | 20-30% | 30-40% | 40-50% | 20-30% | 30-40% | 40-50% |
| GL [Iizuka et al., 2017] | 21.33 | 19.11 | 17.56 | 0.7672 | 0.6823 | 0.5987 | 0.1847 | 0.2535 | 0.3189 | 39.22 | 53.24 | 68.46 |
| CA [Yu et al., 2018] | 20.44 | 18.63 | 17.30 | 0.7652 | 0.6906 | 0.6133 | 0.1948 | 0.2490 | 0.3064 | 30.21 | 40.28 | 53.38 |
| PICNet [Zheng et al., 2019] | 24.44 | 22.32 | 20.71 | 0.8520 | 0.7850 | 0.7119 | 0.1183 | 0.1666 | 0.2245 | 21.62 | 29.59 | 41.60 |
| HiFill [Yi et al., 2020] | 22.54 | 20.15 | 18.48 | 0.7838 | 0.7057 | 0.6193 | 0.1632 | 0.2258 | 0.3053 | 26.89 | 38.40 | 56.24 |
| ICT [Wan et al., 2021] | 24.53 | 22.84 | 21.11 | 0.8599 | **0.7995** | 0.7228 | 0.1045 | 0.1563 | 0.1974 | 17.13 | 22.39 | 28.18 |
| Ours-*Coarse*, Top1 | 24.49 | 22.35 | 20.88 | 0.8449 | 0.7778 | 0.7124 | 0.1052 | 0.1520 | 0.1911 | 17.20 | 17.04 | 17.43 |
| Ours-*Coarse*, Random | 23.49 | 21.39 | 19.95 | 0.8423 | 0.7750 | 0.7102 | 0.1056 | 0.1496 | 0.1970 | 16.70 | 16.58 | 16.85 |
| Ours-*Refine*, Top1 | **25.10** | **22.87** | **21.36** | **0.8640** | 0.7985 | **0.7333** | **0.0871** | 0.1272 | 0.1699 | **15.75** | **15.42** | **15.65** |
| Ours-*Refine*, Random | 24.06 | 21.84 | 20.36 | 0.8613 | 0.7950 | 0.7297 | 0.0872 | **0.1268** | **0.1687** | 15.95 | 15.68 | 16.05 |

Therefore, following the latest two-stage image completion frameworks [Wan et al., 2021, Yi et al., 2020, Yu et al., 2018, 2019], a refinement network is employed to further ensure visual consistency. As shown in Fig. 3 (c), the coarse completion result will first be resized to the original image resolution and recomposed with the original high-quality visible pixels by:

$$x_{comp} = M \odot x_m + (1 - M) \odot \hat{x}, \qquad (15)$$

where $M$ is the initial binary mask with 0 denoting holes, $x_m$ is the masked image and $\hat{x}$ is the corresponding output after resizing. Finally, $x_{comp}$ is passed to a fully convolutional encoder and decoder, where a contextual attention module [Zheng et al., 2022] is applied to copy the high-frequency visible features from the encoder to the decoder. Here, we directly used the latest framework TFill [Zheng et al., 2022] to produce high-quality results, with diverse appearances. The training losses for the refinement network consist of the reconstruction loss ((4) and (5)) and the adversarial loss (7).

## 4 EXPERIMENTS

### 4.1 Experimental Details

*Datasets.* We trained and evaluated our model on three different datasets: **FFHQ** [Karras et al., 2019], a high-quality image dataset of human faces, in which following TFill [Zheng et al., 2022], we selected 60,000 images for training, and 10,000 images for testing; **ImageNet** [Russakovsky et al., 2015], a natural image dataset with various object categories, in which we used original split that 1,281,166 images for training and others for testing; and **Places2** [Zhou et al., 2018], a large scale natural image dataset, in which 8,097,967 images are used for training, and 12,000 images are orderly selected for quantitative evaluation.

*Metrics.* Previous works [Yu et al., 2018, Zheng et al., 2019] have argued that it should *not* be required that the completed output be exactly the same as the original visible image, especially when holes are large. However, for the purpose of quantitative comparison, we report results on various image quality metrics, including traditional peak signal-to-noise ratio (**PSNR**), structure similarity index (**SSIM**), and the latest feature-based learned perceptual image patch similarity (**LPIPS**) [Zhang et al., 2018] and Fréchet Inception Distance (**FID**) [Heusel et al., 2017]. As our model is able to provide multiple solutions for a masked image, we evaluated the top-1 and random of top-*k* results for each quantitative evaluation. Unless otherwise noted, we sampled 10 examples from the top-20 alternative entries.

*Implementation Details.* Our model is trained in three stages: **a)** the code-shared codebook is first trained using the original ground truth image with fixed resolution, *i.e.* 256 × 256. **b)** ours-*Coarse* is then trained for 256 × 256 resolution by inferring the embedded tokens via a highly expressive transformer architecture; and **c)** the ours-*Refined* is finally trained for 512 × 512 resolution. For codebook sizes, we used $K=1024$ for FFHQ, and $K=16384$ for the other datasets.

As codebook training is a separate offline process, our inference pipeline consists of stages **b)** and **c)**. In particular, given a masked image, it will first be downsampled to the fixed resolution, *i.e.* 256 × 256, for diverse content generation. Note that as the refinement network is a fully convolutional network, we can upsample the coarse result to any original resolution, rather than fixing at 512×512 resolution in training. This enables our proposed model to process various images with arbitrary sizes.

### 4.2 Comparison with Existing work

We first performed a thorough comparison of our model to the state-of-the-art methods. Following existing works [Nazeri et al., 2019, Wan et al., 2021], the quantitative evaluation was done on Places2, in which images were degraded by the publicly available free-form masks [Liu et al., 2018]. We report the above-mentioned metrics including PSNR, SSIM, LPIPS, and FID scores. Then, the qualitative results on various datasets are provided for visual comparisons.

*Quantitative Comparison.* We compare our approach to the state-of-the-art methods for free-form image completion in Table 1. Most instantiations of our model outperformed previous state-of-the-art models. These include GL [Iizuka et al., 2017], the first learning-based method for arbitrary regions; CA [Yu et al., 2018] the first method combining learning- and patch-based method; PICNet [Zheng et al., 2019], the first to focus on generating multiple and diverse results; HiFill [Yi et al., 2020], the first method for high-resolution (8K) image completion; ICT [Wan et al., 2021], the latest state-of-the-art work towards generating diverse results using a transformer in discrete space. Without extra bells and whistles, our model outperformed the latest ICT, which also uses a deep transformer architecture to predict the possible discrete tokens for pluralistic image completion. The key difference is that *our model learns an expressive and compositionally flexible codebook in the feature domain, instead* *of using a pre-clustered palette on pixel-level.* Therefore, despite only modeling a shorter sequence distribution, our method can achieve higher image quality after decoding. More interestingly, our framework is quite robust to different mask sizes. Although our paired evaluations on PSNR, SSIM, and LPIPS become gradually worse for larger holes, our FID scores, measuring the dataset-level distribution, remained about the same on different mask ratios, and even better on larger holes. This suggests that while our completed results do not exactly match the corresponding ground truth instances, the diverse solutions fit well to the dataset distribution. We consider this to be appropriate for real image completion, as the dataset ground truth is only a single instance of a whole range of plausible scenes that can give rise to the same unmasked visible region of an image.

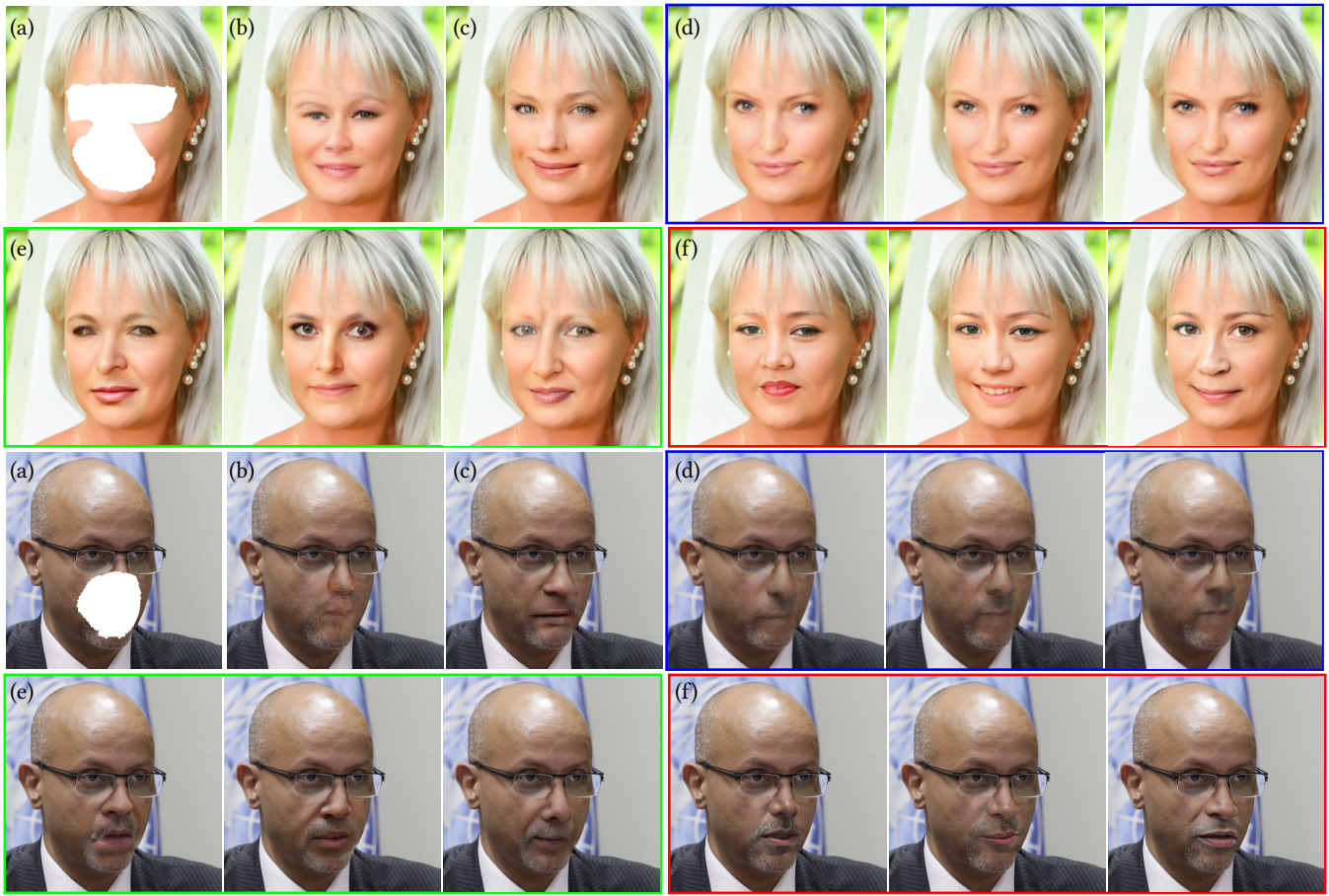

Fig. 7. Comparisons with existing works on FFHQ datasets [Karras et al., 2019]. (a) Masked image. (b) EC [Nazeri et al., 2019] and (c) DeepFillv2 [Yu et al., 2019] generate new contents for missing regions, but with worse image quality and one single solution. (d) PIC [Zheng et al., 2019] provides multiple results, but with limited diversity. (e) While ICT [Wan et al., 2021] improves the diversity, the generated images tend to be of reduced quality. (f) Our model achieves better image quality and larger diversity. More diverse solutions are presented as **animations** in the last term. Best viewed in Adobe Reader. Another 100 examples are provided in the Appendix.

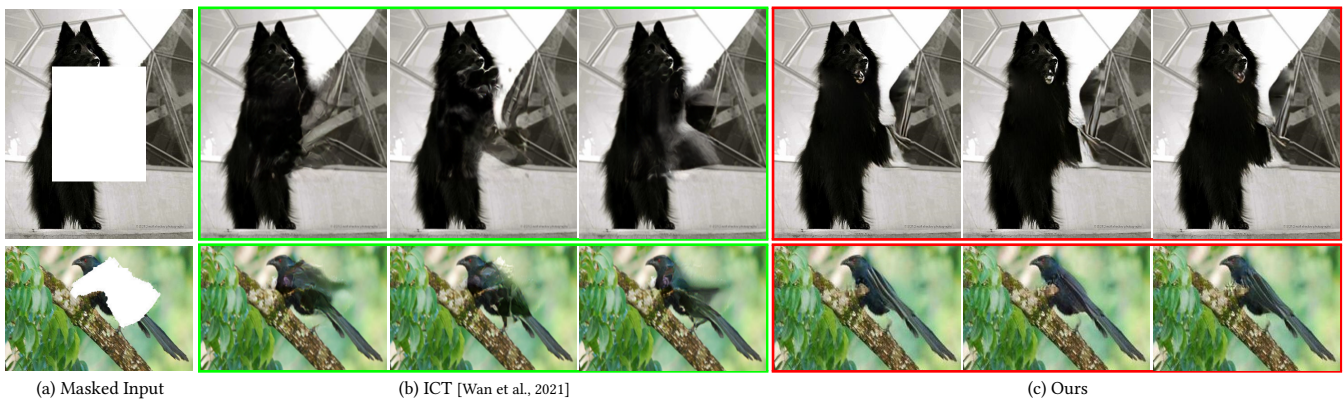

(a) Masked Input          (b) ICT [Wan et al., 2021]          (c) Ours

Fig. 8. Comparisons with existing works on ImageNet datasets [Russakovsky et al., 2015]. While ICT [Wan et al., 2021] provides diverse results, the heavily missed semantic content is hard to be met. In contrast, our model is able to provide some reasonable guess for the large regions. More diverse solutions are presented as **animations** in the last columns of each case. Best viewed in Adobe Reader.

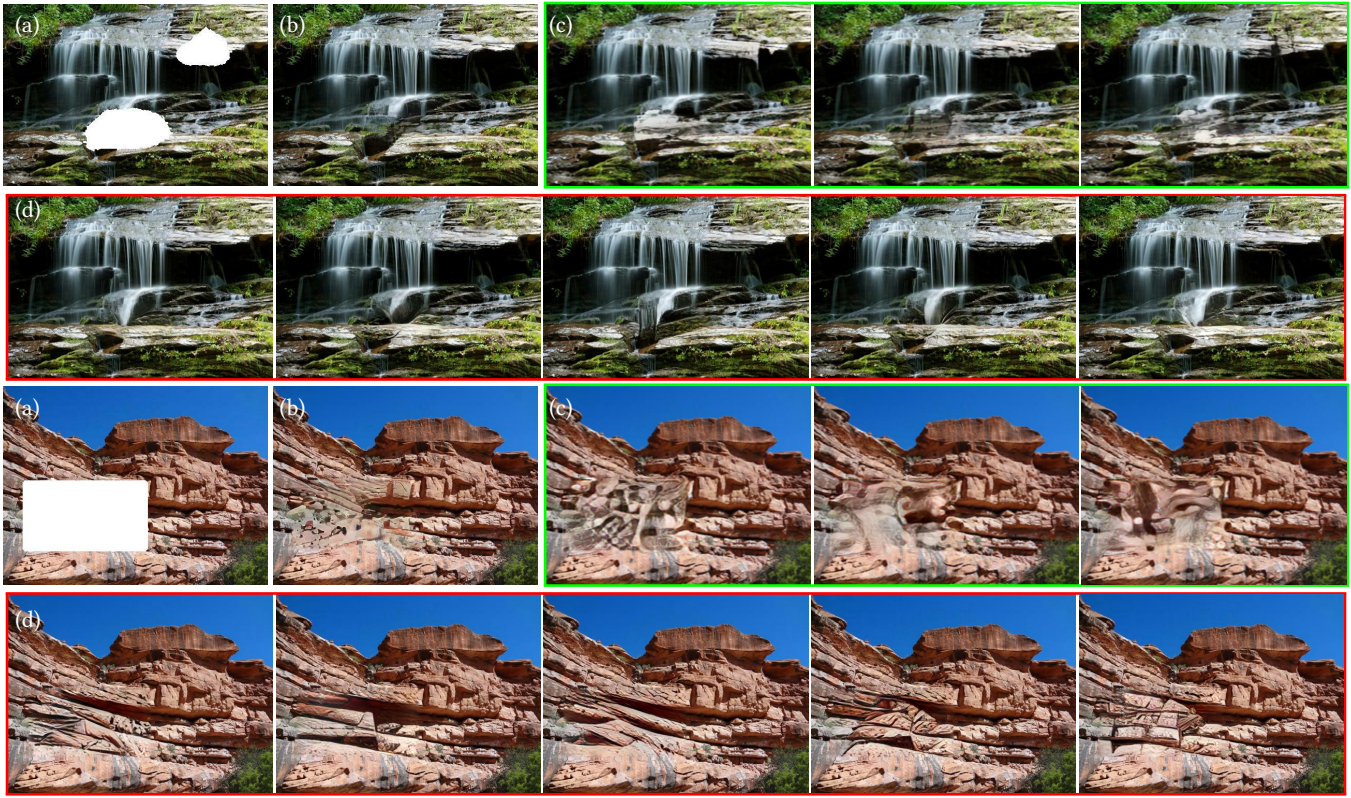

Fig. 9. Comparisons with existing works on Places2 datasets [Zhou et al., 2018]. (a) Masked image. (b) HiFill [Yi et al., 2020] generates reasonable results, but with only one solution. (c) While ICT [Wan et al., 2021] provides multiple solutions, the image quality is worse for large holes. (d) Our method generates multiple and diverse results with high quality. More diverse solutions are presented as **animations** in the final examples. Best viewed in Adobe Reader.

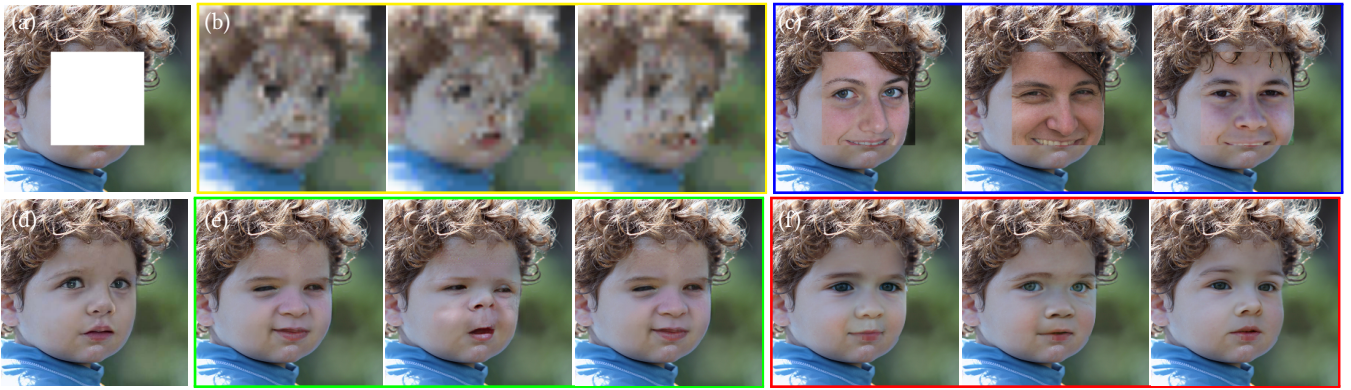

Fig. 10. Comparison of training with different token embedding methods for transformer. (a) Masked input. (b) Results of iGPT [Chen et al., 2020]. (c) Results of VQGAN [Esser et al., 2021]. (d) The single one solution of TFill [Zheng et al., 2022]. (e) Results of the latest state-of-the-art ICT [Wan et al., 2021]. (f) With a better discrete codebook available, our method provides different sizes and colors of eyes and mouth. Note that, all methods employ a transformer architecture of iGPT [Chen et al., 2020], except that the attention module is different in different methods.

Table 2. The effect of different token representations on FFHQ dataset. "Memory" denotes the memory (GB) cost during testing, and "Time" is the average testing time (s) for one center masked image. All approaches employ the transformer architecture to model the global context relationship, except for the different token embedding methods. Following ICT [Wan et al., 2021], all scores are reported on 256 × 256 resolution for a fair comparison.

| Method | LPIPS ↓ | FID ↓ | Memory ↓ | Time ↓ |
|---|---|---|---|---|
| IGPT [Chen et al., 2020] | 0.609 | 148.42 | 3.16 | 26.45 |
| VQGAN [Esser et al., 2021] | 0.226 | 11.92 | **2.36** | 4.29 |
| ICT [Wan et al., 2021] | 0.061 | 4.24 | 3.87 | 152.48 |
| TFill [Zheng et al., 2022] | 0.053 | 3.50 | 3.59 | 0.53 |
| Ours-*Coarse*, Top1 | **0.042** | 2.19 | 3.83 | **0.03** |
| Ours-*Coarse*, Random | 0.044 | **1.53** | 3.83 | **0.03** |

Table 3. The effect of different sampling strategies. Top-$\mathcal{K}$ is the number of candidates. "Autoregressive" sampling needs to predict each token one-by-one via an expensive loop. "One-time" denotes to independently sample all tokens at one time. Here, the LPIPS is for diversity as in [Wan et al., 2021, Zheng et al., 2019], where larger value denotes lager diversity.

| Method | Numbers | LPIPS ↑ | FID ↓ | Time ↓ |
|---|---|---|---|---|
| Autoregressive | Top-1 | - | 5.60 | 3.532 |
| | Top-20 | 0.073 | 5.59 | |
| | Top-40 | 0.097 | 6.53 | |
| | Top-100 | 0.151 | 6.26 | |
| One-time | Top-1 | - | 2.19 | 0.033 |
| | Top-20 | 0.062 | 1.53 | |
| | Top-40 | 0.088 | 1.98 | |
| | Top-100 | 0.124 | 1.77 | |

*Qualitative Comparison.* The qualitative comparisons are visualized in Figs. 7, 8, and 9 for faces, objects, and natural scenes, respectively. Our model achieves good results even under challenging scenarios. In Fig. 7, we can see that the proposed model not only fills in reasonable content with visually realistic appearance, but also provides multiple and diverse choices for different faces. In Fig. 8, we further evaluated our model on more challenging ImageNet dataset, which consists of thousands of different categories of objects, rather than only portrait images in FFHQ dataset. For the latest state-of-the-art ICT [Wan et al., 2021], although it can generate multiple and diverse results, it had some difficulty creating plausible completions for arbitrary animals. In contrast, our proposed model was able to provide multiple results for heavily masked animals, such as the mouth and eyes of the dog. Finally, the comparison is conducted on natural scenes in Fig. 9. While some existing approaches can generate visually reasonable results for background completion, most are geared towards providing only a single result. While ICT [Wan et al., 2021] is able to provide multiple and diverse results, it appears to suffer from reduced quality for large holes. In comparison, our model provides more diverse results, with higher image quality.

### 4.3 Ablation Studies

We ran comprehensive ablation studies to analyze the effectiveness of each key point presented in our model. Results are reported on the FFHQ dataset and shown in Tables 2, 3 and Figs. 10, 11.

*Effect of learned codebook and attention module.* We first investigated the influence of the different token embedding methods in Table 2 and Fig. 10. Here, all methods utilize an IGPT-based [Chen et al., 2020] transformer to predict the tokens. We can see that, IGPT [Chen et al., 2020] and VQGAN [Esser et al., 2021] cannot generate semantically consistent content due to the single-directional attention module that only takes previous information in the scanning order. ICT [Wan et al., 2021] achieved diverse reasonable content through the bidirectional attention module, along with a 3× super resolution network. Although the large diversity is met, the generated image is blurry after large scale upsampling. TFill [Zheng et al., 2022] applied the weighted bidirectional attention module in the transformer, which produces plausible results at high resolution, but only has one solution. Using the same transformer architecture, our models outperformed these state-of-the-art models, even by using only coarse results as shown in Table 2. Compared with TFill, our model trained on discrete space seems to be able to infer more reasonable content. We believe this is because a more compact discrete space is much easier for distribution modeling. Furthermore, the transformer learning involves optimizing a log-likelihood function, instead of seeking a balance in the adversarial function in TFill. Interestingly, our random results also achieved higher image quality than the deterministic result in TFill. Even more surprisingly, our random sampling results led to better FID scores than all other methods (including our top-1 sampling results). This phenomenon suggests that *the distribution inferred by our model is close to the true data distribution*, as our randomly sampled solutions fit well to it, and we are able to avoid out-of-distribution noisy samples.

*Effect of sampling strategy.* Is sequence sampling necessary for image completion? To answer this question, we ran a number of comparisons in Table 3 and Fig. 11. Here, except for different sampling strategies, we used the same trained model for the evaluation. Compared to autoregressive sampling, simultaneously sampling achieves more impressive results in our setting. This is quite surprising. Our conjecture is that the transformer has learned the global context well within the image, and the sampling is appropriately conditioned on the visible regions.

*Effect of sampling numbers.* We also evaluated our model with different numbers of candidates. For this experiment, we first selected the top-$\mathcal{K}$ candidates from the predicted token distribution. We then sample the tokens based on their confidence scores. As can be seen in Table 3, more candidates result in larger diversity, but with worse image quality.

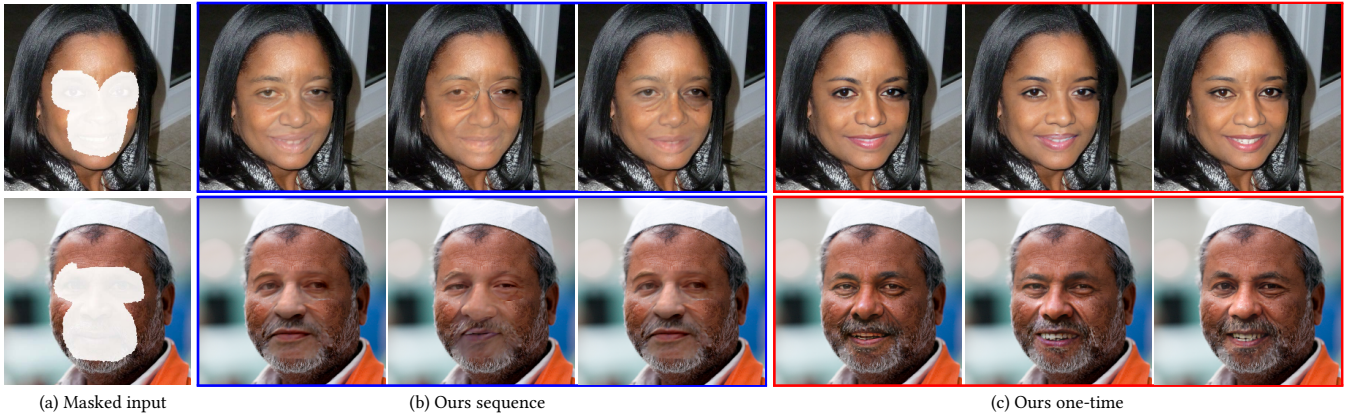

(a) Masked input          (b) Ours sequence          (c) Ours one-time

Fig. 11. Comparison of different sampling strategies during testing. As our training directly predicts all tokens at one time, instead of sequentially depending on the previous scanning line, the sequential generation in our model performs worse than sampling at one time.

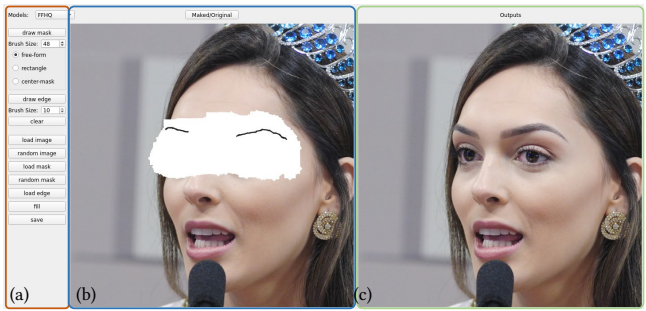

Fig. 12. A screenshot of our interface for free-form image editing. (b) and (c) are the original image and the modified output, respectively. (a) is the control function panel for user input. We can directly mask the target region, and then fill in with multiple and diverse results by clicking the "fill" button. Please refer to the video for this demo.

## 5 APPLICATIONS

Our trained model can be applied to a wide range of applications, including object removal and free-form image editing.

### 5.1 Image Editing Interface

We designed a real-time mask-sketch-based user interface (Fig. 12), enabling image modifications via masks and auxiliary sketches. The control panel (Fig. 12 (a)) consists of some necessary tools such as model selection, image selection, manual input mask and sketch, etc. For a given masked image (Fig. 12 (b)), our system generates completed images with diverse results in real-time on a GPU, by simply clicking the "Fill" button. Users can then select the best result according to their preferences.

### 5.2 Image Editing

With such an interface available, we can now freely edit an image by inputting a mask. The main applications include object removal and more advanced foreground object completion and manipulation.

*Object removal.* The object removal examples are shown in the appendix and video. Our method generally works very well for large object removal by correctly inferring the content based on the partially visible context.

*Free-form image editing.* Instead of filling background pixels into masks for object removal, it is more challenging to generate diverse plausible results for partially visible content. This needs the model to hallucinate new content based on what it has observed, rather than purely completing background textures within an image. As shown in Fig. 14, through masking the mouth of a face, we can synthesize different target expressions. In addition, the model also can generate different shapes for the mountains, after masking the target regions.

### 5.3 Auxiliary Input

Our framework can easily be adapted to include auxiliary input guidance, such as simple user-drawn sketches. Here, we first transform the images in the dataset to sketches using traditional Canny edge detector and the latest learning-based PhotoSketch [Li et al., 2019]. During training, a proportion (we use 40%) of masked images contain corresponding sketches interposed within the masked regions. These are then passed through the encoder and transformer to generate realistic outputs.

We compared our model to the state-of-the-art EdgeConnect [Nazeri et al., 2019] in Fig. 13. Here, we use extracted sketches either from the corresponding image or from the other images. Our method outperforms EdgeConnect by providing better content and consistent appearance. When combining with sketches from other images, the proposed method is able to create multiple new scenes that adapt to the input guidance. In these instances, the diversity is more limited to changing local attributes, while the global structure has been established by the sketches.

Lastly, combining our masking and sketch-based interface, users can freely edit images through masking the target regions and drawing on the corresponding sketches. We show a number of editing examples on faces and natural scenes in Fig. 14, and a qualitative comparison with EdgeConnect [Nazeri et al., 2019] is provided. Edge-Connect cannot provide reasonable content, and exhibits large artifacts on imperfect manual sketches, suggesting that it has difficulty in adapting to arbitrary random manual sketches. Our proposed system mitigates this issue by quantizing manually drawn sketches to the closest tokens, resulting in only small artifacts.

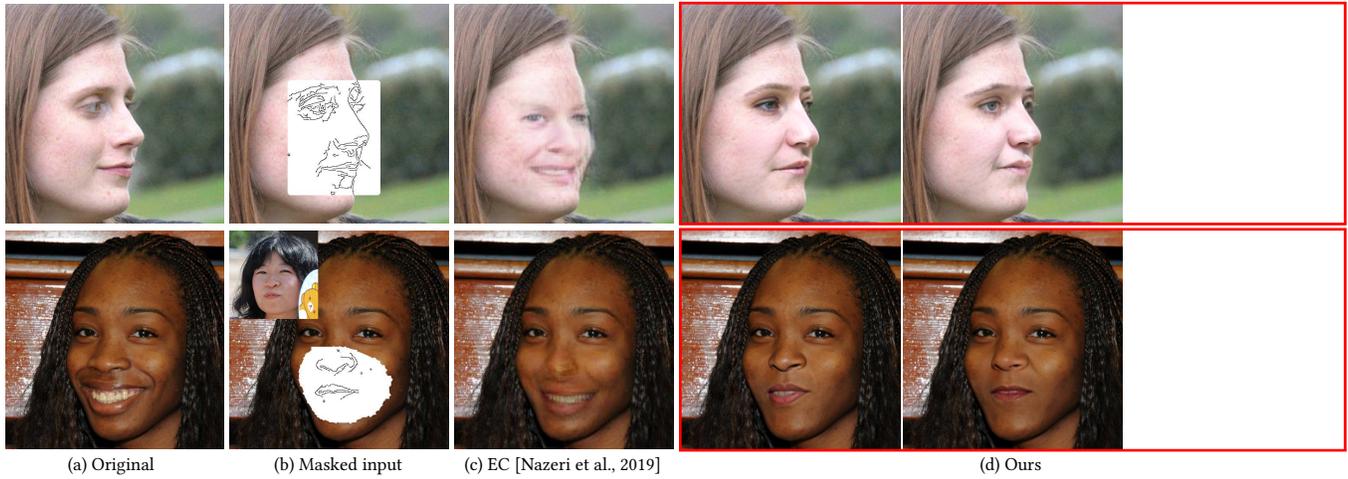

(a) Original      (b) Masked input      (c) EC [Nazeri et al., 2019]      (d) Ours

Fig. 13. Comparisons of image completion given auxiliary edge information. we can use the original canny edge to infer results under challenge scenarios. While the shape is fixed, the details are changed. Furthermore, we can also combine with other edges to recompose a new scene in a cheap way.

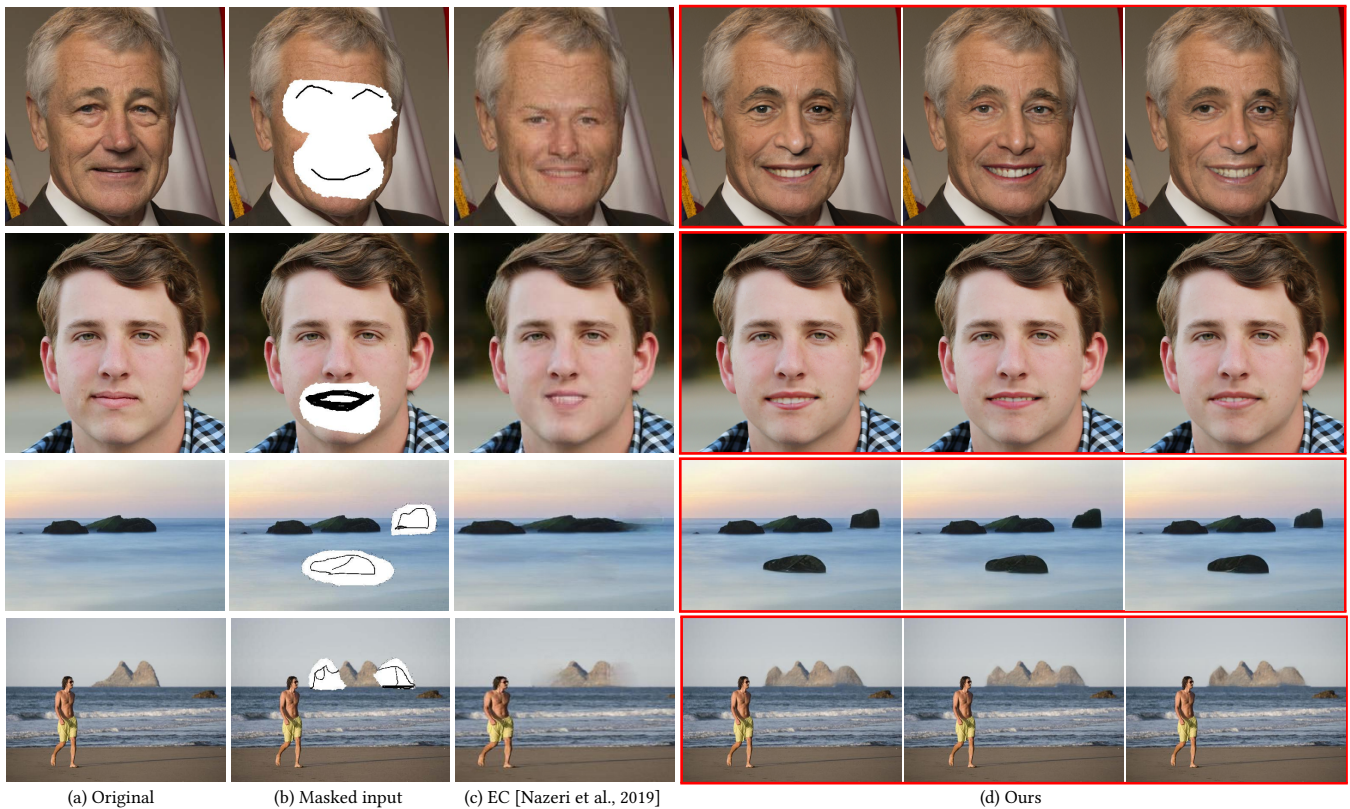

(a) Original      (b) Masked input      (c) EC [Nazeri et al., 2019]      (d) Ours

Fig. 14. Completed results with hand-drawn sketches as auxiliary input information. Our model works well with reasonable hand-drawn sketches. While the shape is guided by the sketch, our method still provides diverse results with different details.

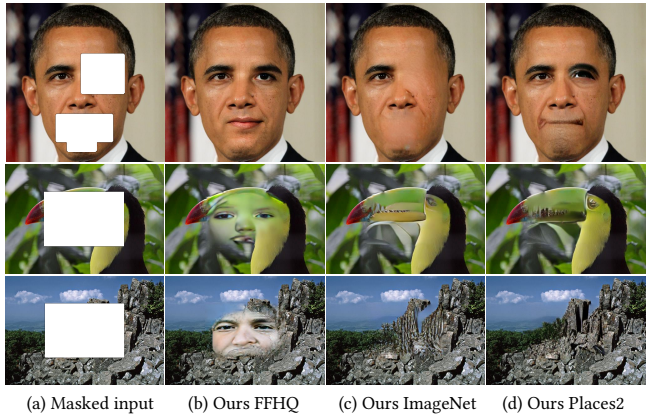

(a) Masked input    (b) Ours FFHQ    (c) Ours ImageNet    (d) Ours Places2

Fig. 15. Examples of models trained on different datasets. We test the model on website images. The special model works well only for a special data type, instead of completing arbitrary images with one model. Note that, as ours FFHQ is trained on face dataset, it aims to provide a face context for various images. In contrast, ours Places2 is trained with a vast number of images, with various scenes. Therefore, it generalizes to various scenarios, while the details are imperfect.

## 6 CONCLUSION

In this paper, we have presented a novel framework for pluralistic image completion that produces multiple and diverse plausible results for a single masked image. We have shown that, by learning a compact and expressive token representation, it is easier to tame a transformer to infer the correct tokens for missing regions. The diverse plausible results are achieved by directly inferring and independently sampling from token distributions in discrete space, bypassing the need for slow autoregressive computation. The appearance is further refined using a fully convolutional network. Through comprehensive experiments and thorough ablation studies, we have demonstrated that our approach can generate much more realistic results with larger diversity than existing works. Finally, we have also shown the application of our system to free-form image editing, such enabling hand-drawn sketch guidance for image completion.

*Limitations.* Although the proposed model is able to provide diverse plausible results for various images that are degraded by random free-form masks, we need to train different models for different data types, *e.g.* faces, animals, objects, buildings and natural scenes. The trained model can not be generalized to arbitrary images in the real world. In Fig. 15, we evaluated the trained model on natural images from websites. As can be seen, the model worked well when images contain the general content found in the dataset, but failed when tested on out-of-distribution images. Therefore, a long-term goal is to train a general codebook, and then tame a network to generate reasonable content for any arbitrary images in the real world. Note that the model trained on larger Places2 dataset is able to achieve reasonable results for faces, objects, and natural scenes, although details are imperfect. This suggests that it is possible to realize such a longer-term goal.

## REFERENCES

C. Ballester, M. Bertalmio, V. Caselles, G. Sapiro, and J. Verdera. Filling-in by joint interpolation of vector fields and gray levels. *IEEE transactions on image processing*, 10(8):1200–1211, 2001.

C. Barnes, E. Shechtman, A. Finkelstein, and D. B. Goldman. Patchmatch: A randomized correspondence algorithm for structural image editing. *ACM Transactions on Graphics (ToG)*, 28:24, 2009.

M. Bertalmio, G. Sapiro, V. Caselles, and C. Ballester. Image inpainting. In *Proceedings of the 27th annual conference on Computer graphics and interactive techniques*, pages 417–424. ACM Press/Addison-Wesley Publishing Co., 2000.

M. Bertalmio, L. Vese, G. Sapiro, and S. Osher. Simultaneous structure and texture image inpainting. *IEEE transactions on image processing*, 12(8):882–889, 2003.

T. B. Brown, B. Mann, N. Ryder, M. Subbiah, J. Kaplan, P. Dhariwal, A. Neelakantan, P. Shyam, G. Sastry, A. Askell, S. Agarwal, A. Herbert-Voss, G. Krueger, T. Henighan, R. Child, A. Ramesh, D. M. Ziegler, J. Wu, C. Winter, C. Hesse, M. Chen, E. Sigler, M. Litwin, S. Gray, B. Chess, J. Clark, C. Berner, S. McCandlish, A. Radford, I. Sutskever, and D. Amodei. Language models are few-shot learners. 2020.

M. Chen, A. Radford, R. Child, J. Wu, H. Jun, D. Luan, and I. Sutskever. Generative pretraining from pixels. In *Proceedings of the International Conference on Machine Learning*, pages 1691–1703. PMLR, 2020.

A. Criminisi, P. Perez, and K. Toyama. Object removal by exemplar-based inpainting. In *Proceedings of the IEEE Computer Society Conference on Computer Vision and Pattern Recognition.*, volume 2, pages II–II. IEEE, 2003.

A. Criminisi, P. Pérez, and K. Toyama. Region filling and object removal by exemplar-based image inpainting. *IEEE Transactions on image processing*, 13(9):1200–1212, 2004.

J. Devlin, M.-W. Chang, K. Lee, and K. Toutanova. Bert: Pre-training of deep bidirectional transformers for language understanding. *arXiv preprint arXiv:1810.04805*, 2018.

P. Esser, R. Rombach, and B. Ommer. Taming transformers for high-resolution image synthesis. In *Proceedings of the IEEE/CVF Conference on Computer Vision and Pattern Recognition (CVPR)*, pages 12873–12883, 2021.

I. Goodfellow, J. Pouget-Abadie, M. Mirza, B. Xu, D. Warde-Farley, S. Ozair, A. Courville, and Y. Bengio. Generative adversarial nets. In *Proceedings of the Advances in neural information processing systems*, pages 2672–2680, 2014.

J. Hays and A. A. Efros. Scene completion using millions of photographs. *ACM Transactions on Graphics (TOG)*, 26:4, 2007.

M. Heusel, H. Ramsauer, T. Unterthiner, B. Nessler, and S. Hochreiter. Gans trained by a two time-scale update rule converge to a local nash equilibrium. In *Proceedings of the 31st International Conference on neural information processing systems*, pages 6626–6637, 2017.

S. Iizuka, E. Simo-Serra, and H. Ishikawa. Globally and locally consistent image completion. *ACM Transactions on Graphics (TOG)*, 36(4):107, 2017.

J. Jia and C.-K. Tang. Inference of segmented color and texture description by tensor voting. *IEEE Transactions on Pattern Analysis and Machine Intelligence*, 26(6):771–786, 2004.

Y. Jo and J. Park. Sc-fegan: Face editing generative adversarial network with user's sketch and color. In *Proceedings of the IEEE/CVF International Conference on Computer Vision (ICCV)*, October 2019.

T. Karras, S. Laine, and T. Aila. A style-based generator architecture for generative adversarial networks. In *Proceedings of the IEEE/CVF Conference on Computer Vision and Pattern Recognition (CVPR)*, pages 4401–4410, 2019.

R. Köhler, C. Schuler, B. Schölkopf, and S. Harmeling. Mask-specific inpainting with deep neural networks. In *Proceedings of the German Conference on Pattern Recognition*, pages 523–534. Springer, 2014.

Y. LeCun, L. Bottou, Y. Bengio, and P. Haffner. Gradient-based learning applied to document recognition. *Proceedings of the IEEE*, 86(11):2278–2324, 1998.

A. Levin, A. Zomet, and Y. Weiss. Learning how to inpaint from global image statistics. In *Proceedings of the IEEE International Conference on Computer Vision (ICCV)*, volume 1, pages 305–312. IEEE, 2003.

J. Li, N. Wang, L. Zhang, B. Du, and D. Tao. Recurrent feature reasoning for image inpainting. In *Proceedings of the IEEE/CVF Conference on Computer Vision and Pattern Recognition (CVPR)*, June 2020.

M. Li, Z. Lin, R. Mech, E. Yumer, and D. Ramanan. Photo-sketching: Inferring contour drawings from images. In *2019 IEEE Winter Conference on Applications of Computer Vision (WACV)*, pages 1403–1412. IEEE, 2019.

Y. Li, S. Liu, J. Yang, and M.-H. Yang. Generative face completion. In *Proceedings of the IEEE Conference on Computer Vision and Pattern Recognition (CVPR)*, pages 5892–5900. IEEE, 2017.

L. Liao, J. Xiao, Z. Wang, C.-W. Lin, and S. Satoh. Image inpainting guided by coherence priors of semantics and textures. In *Proceedings of the IEEE/CVF Conference on Computer Vision and Pattern Recognition (CVPR)*, pages 6539–6548, June 2021.

G. Liu, F. A. Reda, K. J. Shih, T.-C. Wang, A. Tao, and B. Catanzaro. Image inpainting for irregular holes using partial convolutions. In *Proceedings of the European Conference on Computer Vision (ECCV)*, September 2018.

H. Liu, Z. Wan, W. Huang, Y. Song, X. Han, and J. Liao. Pd-gan: Probabilistic diverse gan for image inpainting. In *Proceedings of the IEEE/CVF Conference on Computer*

*Vision and Pattern Recognition (CVPR)*, pages 9371–9381, June 2021.

T. Miyato, T. Kataoka, M. Koyama, and Y. Yoshida. Spectral normalization for generative adversarial networks. In *Proceedings of the International Conference on Learning Representations*, 2018.

K. Nazeri, E. Ng, T. Joseph, F. Qureshi, and M. Ebrahimi. Edgeconnect: Structure guided image inpainting using edge prediction. In *Proceedings of the IEEE/CVF IEEE International Conference on Computer Vision (ICCV) Workshops*, Oct 2019.

D. Pathak, P. Krahenbuhl, J. Donahue, T. Darrell, and A. A. Efros. Context encoders: Feature learning by inpainting. In *Proceedings of the IEEE Conference on Computer Vision and Pattern Recognition (CVPR)*, pages 2536–2544. IEEE, 2016.

J. Peng, D. Liu, S. Xu, and H. Li. Generating diverse structure for image inpainting with hierarchical vq-vae. In *Proceedings of the IEEE/CVF Conference on Computer Vision and Pattern Recognition (CVPR)*, pages 10775–10784, 2021.

T. Portenier, Q. Hu, A. Szabo, S. A. Bigdeli, P. Favaro, and M. Zwicker. Faceshop: Deep sketch-based face image editing. *ACM Transactions on Graphics (TOG)*, 37(4):99, 2018.

A. Radford and K. Narasimhan. Improving language understanding by generative pre-training. 2018.

A. Radford, J. Wu, R. Child, D. Luan, D. Amodei, and I. Sutskever. Language models are unsupervised multitask learners. 2019.

A. Razavi, A. van den Oord, and O. Vinyals. Generating diverse high-fidelity images with vq-vae-2. In *Proceedings of the 33rd International Conference on Neural Information Processing Systems*, pages 14866–14876, 2019.

J. S. Ren, L. Xu, Q. Yan, and W. Sun. Shepard convolutional neural networks. In *Proceedings of the Advances in Neural Information Processing Systems*, pages 901–909, 2015.

O. Russakovsky, J. Deng, H. Su, J. Krause, S. Satheesh, S. Ma, Z. Huang, A. Karpathy, A. Khosla, M. Bernstein, et al. Imagenet large scale visual recognition challenge. *International Journal of Computer Vision*, 115(3):211–252, 2015.

Y. Song, C. Yang, Z. Lin, X. Liu, Q. Huang, H. Li, and C. Jay. Contextual-based image inpainting: Infer, match, and translate. In *Proceedings of the European Conference on Computer Vision (ECCV)*, pages 3–19, 2018a.

Y. Song, C. Yang, Y. Shen, P. Wang, Q. Huang, and C.-C. J. Kuo. Spg-net: Segmentation prediction and guidance network for image inpainting. In *Proceedings of the British Machine Vision Conference (BMVC)*, September 2018b.

M. Suin, K. Purohit, and A. N. Rajagopalan. Distillation-guided image inpainting. In *Proceedings of the IEEE/CVF International Conference on Computer Vision (ICCV)*, pages 2481–2490, October 2021.

A. Torralba, R. Fergus, and W. T. Freeman. 80 million tiny images: A large data set for nonparametric object and scene recognition. *IEEE transactions on pattern analysis and machine intelligence*, 30(11):1958–1970, 2008.

A. van den Oord, O. Vinyals, and K. Kavukcuoglu. Neural discrete representation learning. In *Proceedings of the 31st International Conference on Neural Information Processing Systems*, pages 6309–6318, 2017.

A. Vaswani, N. Shazeer, N. Parmar, J. Uszkoreit, L. Jones, A. N. Gomez, L. u. Kaiser, and I. Polosukhin. Attention is all you need. In I. Guyon, U. V. Luxburg, S. Bengio, H. Wallach, R. Fergus, S. Vishwanathan, and R. Garnett, editors, *Advances in Neural Information Processing Systems*, volume 30. Curran Associates, Inc., 2017.

Z. Wan, J. Zhang, D. Chen, and J. Liao. High-fidelity pluralistic image completion with transformers. In *Proceedings of the IEEE/CVF International Conference on Computer Vision (ICCV)*, pages 4692–4701, October 2021.

Z. Yan, X. Li, M. Li, W. Zuo, and S. Shan. Shift-net: Image inpainting via deep feature rearrangement. In *Proceedings of the European Conference on Computer Vision (ECCV)*, September 2018.

C. Yang, X. Lu, Z. Lin, E. Shechtman, O. Wang, and H. Li. High-resolution image inpainting using multi-scale neural patch synthesis. In *Proceedings of the IEEE/CVF Conference on Computer Vision and Pattern Recognition (CVPR)*, volume 1, page 3, 2017.

Z. Yi, Q. Tang, S. Azizi, D. Jang, and Z. Xu. Contextual residual aggregation for ultra high-resolution image inpainting. In *Proceedings of the IEEE/CVF Conference on Computer Vision and Pattern Recognition (CVPR)*, pages 7508–7517, 2020.

J. Yu, Z. Lin, J. Yang, X. Shen, X. Lu, and T. S. Huang. Generative image inpainting with contextual attention. In *Proceedings of the IEEE Conference on Computer Vision and Pattern Recognition (CVPR)*, pages 5505–5514, 2018.

J. Yu, Z. Lin, J. Yang, X. Shen, X. Lu, and T. S. Huang. Free-form image inpainting with gated convolution. In *Proceedings of the IEEE International Conference on Computer Vision (ICCV)*, pages 4471–4480, 2019.

Y. Yu, F. Zhan, R. Wu, J. Pan, K. Cui, S. Lu, F. Ma, X. Xie, and C. Miao. Diverse image inpainting with bidirectional and autoregressive transformers. In *Proceedings of the 29th ACM International Conference on Multimedia*, 2021.

Y. Zeng, Z. Lin, J. Yang, J. Zhang, E. Shechtman, and H. Lu. High-resolution image inpainting with iterative confidence feedback and guided upsampling. In *Proceedings of the European Conference on Computer Vision (ECCV)*, pages 1–17. Springer, 2020.

Y. Zeng, Z. Lin, H. Lu, and V. M. Patel. Cr-fill: Generative image inpainting with auxiliary contextual reconstruction. In *Proceedings of the IEEE/CVF International Conference on Computer Vision (ICCV)*, pages 14164–14173, October 2021.

R. Zhang, P. Isola, A. A. Efros, E. Shechtman, and O. Wang. The unreasonable effectiveness of deep features as a perceptual metric. In *Proceedings of the IEEE conference on computer vision and pattern recognition (CVPR)*, pages 586–595, 2018.

L. Zhao, Q. Mo, S. Lin, Z. Wang, Z. Zuo, H. Chen, W. Xing, and D. Lu. Uctgan: Diverse image inpainting based on unsupervised cross-space translation. In *Proceedings of the IEEE/CVF Conference on Computer Vision and Pattern Recognition (CVPR)*, June 2020.

C. Zheng, T.-J. Cham, and J. Cai. Pluralistic image completion. In *Proceedings of the IEEE/CVF Conference on Computer Vision and Pattern Recognition (CVPR)*, June 2019.

C. Zheng, T.-J. Cham, and J. Cai. Pluralistic free-form image completion. *International Journal of Computer Vision*, 129(10):2786–2805, 2021a.

C. Zheng, D.-S. Dao, G. Song, T.-J. Cham, and J. Cai. Visiting the invisible: Layer-by-layer completed scene decomposition. *International Journal of Computer Vision*, 129 (12):3195–3215, 2021b.

C. Zheng, T.-J. Cham, J. Cai, and D. Phung. Bridging global context interactions for high-fidelity image completion. In *Proceedings of the IEEE/CVF Conference on Computer Vision and Pattern Recognition (CVPR)*, 2022.

B. Zhou, A. Lapedriza, A. Khosla, A. Oliva, and A. Torralba. Places: A 10 million image database for scene recognition. *IEEE transactions on pattern analysis and machine intelligence*, 40(6):1452–1464, 2018.

# Supplementary Material: High-quality Pluralistic Image Completion via Code Sharing

The supplementary materials are organized as follows:

(1) A video to illuminate our work and interface;

(2) More results for free-form image completion on FFHQ dataset. We directly show 1-100 index from FFHQ without curated selection.

(3) Results for object removal.

## A OBJECT REMOVAL

Please see results from Fig A.1.

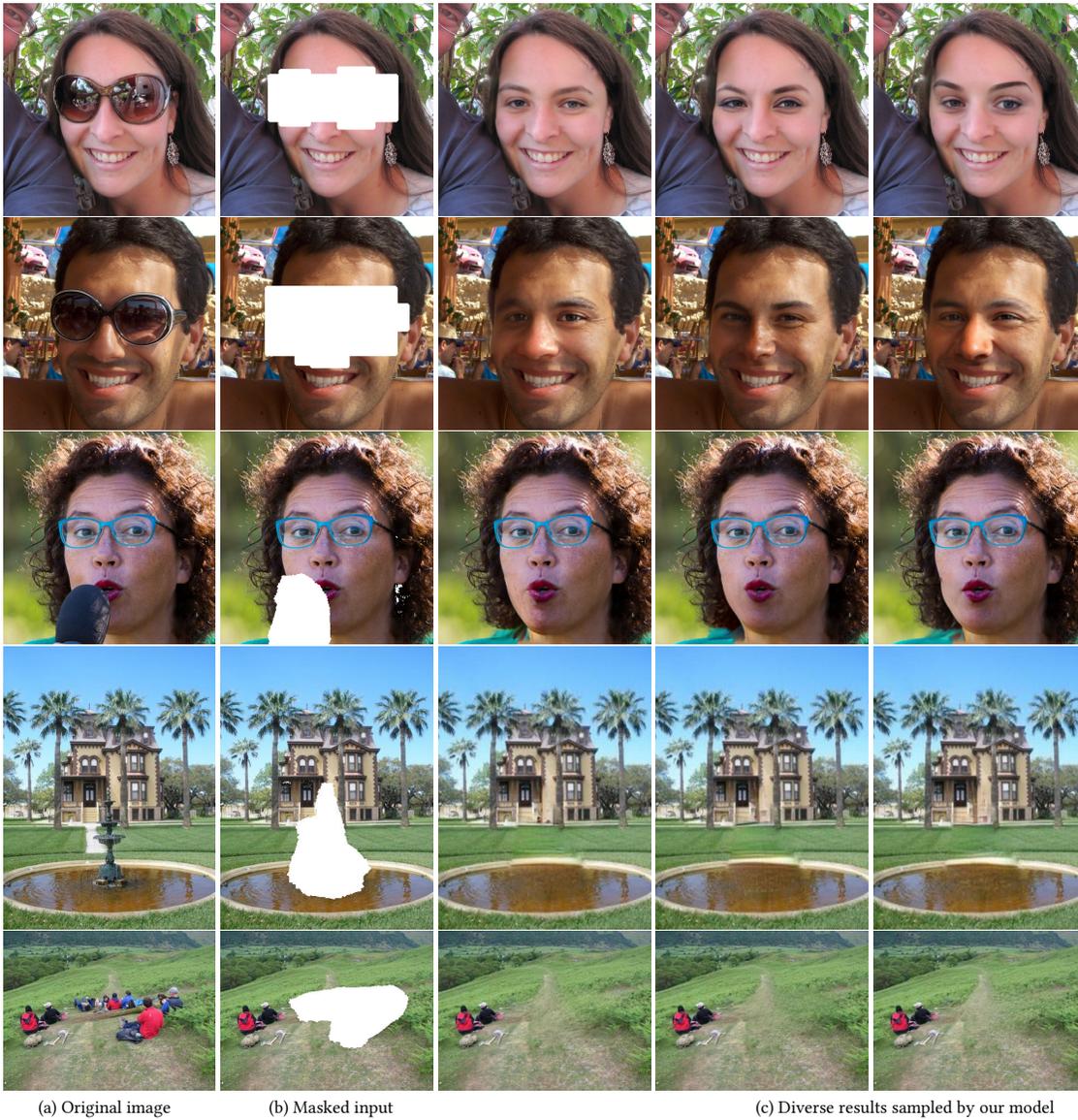

(a) Original image     (b) Masked input             (c) Diverse results sampled by our model

Fig. A.1. Examples of object removal by our approach on faces and natural scenes. The last column shows diverse results as **animations**. It will be more obvious to capture the difference between different solutions. If foreground objects are fully masked, our method performs to fill into with background contents, due to it only captures the background visible information.